\documentclass{article}
\usepackage{spconf,amsmath,graphicx}
\usepackage{fancyhdr}

\fancypagestyle{firstpage}
{
    \fancyhead[L]{Published as a conference paper at ICASSP 2022}    
    \fancyhead[R]{}
}

\title{Personalized PageRank Graph Attention Networks}
\name{Julie Choi}
\address{julichoi@amazon.com}
\begin{document}
%
\maketitle
\thispagestyle{firstpage}

\begin{abstract}
There has been a rising interest in graph neural networks (GNNs) for representation learning over the past few years. GNNs provide a general and efficient framework to learn from graph-structured data. However, GNNs typically only use the information of a very limited neighborhood for each node to avoid over-smoothing. A larger neighborhood would be desirable to provide the model with more information. In this work, we incorporate the limit distribution of Personalized PageRank (PPR) into graph attention networks (GATs) to reflect the larger neighbor information without introducing over-smoothing. Intuitively, message aggregation based on Personalized PageRank corresponds to infinitely many neighborhood aggregation layers. We show that our models outperform a variety of baseline models for four widely used benchmark datasets. Our implementation is publicly available online. \footnote{https://github.com/juliechoi12/pprgat}
\end{abstract}

\section{Introduction}
\label{sec:intro}



Graph Neural Networks (GNNs) \cite{gori2005new,scarselli2008graph} have proved to be an effective representation learning framework for graph-structured data. GNNs have achieved state-of-the-art performance on many practical predictive tasks, such as node classification, link prediction and graph classification. A GNN can be viewed as a message-passing network \cite{gilmer2017neural}, where each node iteratively updates its state by interacting with its neighbors. GNN variants mostly differ in how each node aggregates the representations of its neighbors and combines them with its own representation. Several methods have been proposed to improve the basic neighborhood aggregation scheme such as attention mechanisms \cite{kearnes2016molecular,gat}, random walks \cite{pinsage,li2018deeper}, edge features \cite{gilmer2017neural} and making it more scalable on large graphs \cite{pinsage,fastgcn}. However, all these methods only use the information of a very limited neighborhood for each node to avoid over-smoothing \cite{li2018deeper}. A larger neighborhood would be needed to provide the model with more information.


In this work, we incorporate the limit distribution of Personalized PageRank (PPR) \cite{page1999pagerank} into GNNs to reflect the larger neighbor information without introducing over-smoothing. Message aggregation based on PPR corresponds to infinitely many neighborhood aggregation layers. However, computing the PPR distribution can be challenging for large graphs \cite{appnp,pprgo}. Our approach uses an approximate PPR matrix, which approximates the PPR matrix with a sparse matrix in a scalable and distributed way. Also, to fully leverage the expressive power of GNNs, we incorporate this PPR matrix into graph attention networks (GATs) \cite{gat}. GAT generalizes the standard averaging or max-pooling of neighbors \cite{kipf2017semi,graphsage}, by allowing every node to compute a weighted average of its neighbors according to the neighbor's importance. We call our model PPRGAT (Personalized PageRank Graph Attention Network).


To incorporate the PPR matrix into GAT, we consider two versions. First, we concatenate the PPR matrix information to the node features $x_i, x_j$ to compute the attention score between node $i$ and node $j$ if there is an edge between node $i$ and node $j$. In this version, we use the original adjacency matrix, and the neighbors are determined by this adjacency matrix as in the standard GATs \cite{gat}. Second, we replace the original adjacency matrix with the sparse approximate PPR matrix. In this version, only the top k elements of $i$'s row of the approximate PPR matrix are considered the neighbors of node $i$. We show that our models outperform a variety of baseline models across all datasets used for our experiments.

\section{Related work}
In this section, we first introduce our notation and then review prior work related to our work. Let $G = (V,E)$ denote a graph with a set of nodes $V = \{v_1, \cdots ,v_N\}$, connected by a set of edges $E \subseteq V \times V$. Node features are organized in a compact matrix $X \in R^{N \times D}$  with each row representing the feature vector of one node, where $N$ is the number of nodes and $D$ is the dimension of the features. Let $A \in R^{N \times N}$ denote the adjacent matrix that describes graph structure of $G: A_{ij} = 1$ if there is an edge $e_{ij}$ from node $i$ to node $j$, and $0$ otherwise. By adding a self-loop to each node, we have $\tilde{A} = A + I_{N}$ to denote the adjacency matrix of the augmented graph, where $I_N \in R^{N \times N}$ is an identity matrix.


 
\subsection{Neighbor Aggregation Methods}
Most of the graph learning algorithms are based on a neighbor aggregation mechanism. The basic idea is to learn a parameter-sharing aggregator, which takes feature vector $x_i$ of node $i$ and its neighbors’ feature vectors to output a new feature vector for the node $i$. The aggregator function defines how to aggregate the low level features of a certain node and its neighbors to generate high level feature representations. The popular Graph Convolution Networks (GCNs) \cite{kipf2017semi} fall into the category of neighbor aggregation. For a node classification task with a 2-layer GCN, its encoder function can be expressed as
\begin{equation} \label{eq:encoder}
f(X,A,W) = \text{softmax} \left( \hat{A}\sigma (\hat{A} XW^{(0)} )W^{(1)} \right),
\end{equation}
where $\hat{A} = \tilde{D}^{- \frac{1}{2}} \tilde{A} \tilde{D}^{- \frac{1}{2}}$, $D_{ii} = \sum_{j} \tilde{A}_{ij}$, $W^{(\cdot)}$s are the learnable parameters of GCNs, and $\sigma(\cdot)$ is a nonlinearity function. Specifically, the aggregator of GCNs can be expressed as
\begin{equation} \label{eq:gcn-agg}
h_i^{(l+1)} = \sigma\left(\sum_{j\in \mathcal{N}_i} \tilde{A}_{ij} h_j^{(l)} W^{(l)} \right),
\end{equation}
where $h_j^{(l)}$ is the hidden representation of node $j$ at layer $l$
, $h^{(0)} = X$, and $\mathcal{N}_i$ denotes the set of all the neighbors of node $i$, including itself.


\subsection{Graph Attention Networks}
\label{ssec:gat}
Recently, attention networks have achieved state-of-the-art results in many tasks \cite{xu2015show}. By using learnable weights on each input, the attention mechanism determines how much attention to give to each input. GATs \cite{gat} utilize an attention-based aggregator to generate attention coefficients over all neighbors of a node for feature aggregation. In particular, the aggregator function of GATs is
\begin{equation} \label{eq:gat-agg}
h_i^{(l+1)} = \sigma\left(\sum_{j\in \mathcal{N}_i} \alpha_{ij}^{(l)} h_j^{(l)} W^{(l)} \right)
\end{equation}
This is similar to that of GCNs, Equation \eqref{eq:encoder}, except that $\alpha_{ij}^{(l)}$ is the attention coefficient of edge $e_{ij}$
at layer $l$, assigned by an attention function rather than by a predefined $\tilde{A}$. To increase the capacity of attention mechanism, GATs further exploit multi-head attentions for feature aggregation. The learned attention coefficient can be viewed as an importance score of an edge.

In GAT, a scoring function $e: R^d \times R^d \rightarrow R$ computes a score for every edge $(j, i)$, which indicates the importance of the features of the neighbor $j$ to the node $i$:
\begin{equation} \label{eq:gat-e}
e(h_i,h_j)=\text{LeakyReLU} (a^{T} \cdot [W h_i || W h_j])
\end{equation}
where $a \in R^{2d'}, W \in R^{d' \times d}$ are learned, and $||$ denotes vector concatenation. These attention scores are normalized across all neighbors $j \in \mathcal{N}_i$ using softmax:
\begin{equation} \label{eq:gat-attention}
\alpha_{ij} = \text{softmax}_j (e(h_i, h_j)) = \frac{\exp(e(h_i, h_j))}{\sum_{j'\in \mathcal{N}_i} \exp(e(h_i, h_{j'})})
\end{equation}


Later, \cite{gatv2} pointed out that the ranking of attention coefficients in Equation (\ref{eq:gat-attention}) is global for all nodes. \cite{gatv2} proposed GATv2, which modifies Equation (\ref{eq:gat-e}) to
\begin{equation} \label{eq:gatv2-e}
e(h_i,h_j)=a^{T} \text{LeakyReLU} (W \cdot [h_i ||h_j])
\end{equation}
and showed that it can yield dynamic attention using this simple modification.

In our work, we implement our PPRGAT for these two versions, GAT \cite{gat} and GATv2 \cite{gatv2}. We will call the second version (PPRGAT applied to GATv2) as PPRGATv2 if the distinguishment is needed. We show that PPRGAT and PPRGATv2 outperform the baselines.

\subsection{Personalized PageRank and GNNs}
\label{ssec:PPR-GNN}
Despite their success of GNNs, over-smoothing is a common issue faced by GNNs \cite{li2018deeper}. That is, the representations of the graph nodes of different classes would become indistinguishable when stacking multiple layers, which hurts the model performance. Due to this reason, GNNs typically only use the information of a very limited neighborhood for each node. In order to address this issue, there have been efforts to incorporate PPR into GNNs recently. \cite{appnp} proposes "Predict then Propagate" methods: It first predicts the node representations using a simple MLP, and then aggregates these node representations using PPR limit distributions. \cite{pprgo} uses a similar approach, except that it pre-computes the limit distributions and uses only top k important neighbors of each node for scalability.

\section{Proposed Method: PPRGAT}
\label{sec:method}

In this section, we introduce our new model, PPRGAT. Our work is inspired by GAT (Section~\ref{ssec:gat}) and PPR (Section~\ref{ssec:PPR-GNN}). We incorporate PPR limit distribution into the attention coefficients in GAT layers. In this way, GNNs can be aware of the whole graph structure and learn more efficient attention weights.

\subsection{Approximate PPR Distribution as Sparse Matrix}
\label{ssec:ppr-matrix}

We consider the PPR matrix defined by 
\begin{equation} \label{eq:ppr-matrix}
\Pi^{\text{ppr}} = \alpha \left( I_n - (1-\alpha) D^{-1}A\right)^{-1},
\end{equation}
where is a teleport probability. Each row $\pi(i) := \Pi_{i,:}^{\text{ppr}}$ is equal to the PPR vector of node $i$. We are interested in efficient and scalable algorithms for computing (an approximation of) PPR. Random walk sampling \cite{graphsage} is one such approximation technique. While it's simple to implement, in order to guarantee at most $\epsilon$ absolute error with probability of $1 - \frac{1}{n}$, we need $O\left(\frac{\log n}{\epsilon^2}\right)$ random walks.

For this work, we adopt the approach by \cite{andersen2006local}. It offers a good balance of scalability and approximation guarantees. When the graph is strongly connected, $\pi(i)$ is non-zero for all nodes. However, we can still obtain a good approximation by truncating small elements to zero since most of the probability mass in the PPR vectors $\pi(i)$ is localized on a small number of nodes \cite{andersen2006local,nassar2015strong}. Thus, we can approximate $\pi(i)$ with a sparse vector and in turn approximate $\Pi^{\text{ppr}}$ with a sparse matrix $\Pi^{\epsilon}$.


We additionally truncate $\Pi^{\epsilon}$ to contain only the top $k$ largest entries for each row $\pi(i)$. We call it $\Pi^{\epsilon, k}$. Note that this computation can be parallelized for each node $i$. Furthermore, we only need to pre-compute $\Pi^{\epsilon, k}$ once and reuse it during the training.

\subsection{GAT layer with PPR matrix}
\label{ssec:gat-ppr}
To incorporate the PPR matrix information into the GAT layer, we modify Equation~(\ref{eq:gat-e}) to

\begin{equation} \label{eq:pprgat-e}
e(h_i,h_j)=\text{LeakyReLU} (a^{T} \cdot [W h_i \| W h_j \| \Pi^{\epsilon, k}_{ij}]).
\end{equation}

Using this simple modification, we can naturally incorporate the global graph structure into the local GAT layer. We apply the same approach to GATv2 \cite{gatv2} as well. For this, Equation (\ref{eq:gatv2-e}) is modified to
\begin{equation} \label{eq:pprgatv2-e}
e(h_i,h_j)=a^{T} \text{LeakyReLU} (W \cdot [h_i \|h_j\| \Pi^{\epsilon, k}_{ij}])
\end{equation}
We call this model PPRGATv2.

\begin{figure}[h]

\begin{center}

\includegraphics[width=1\linewidth]{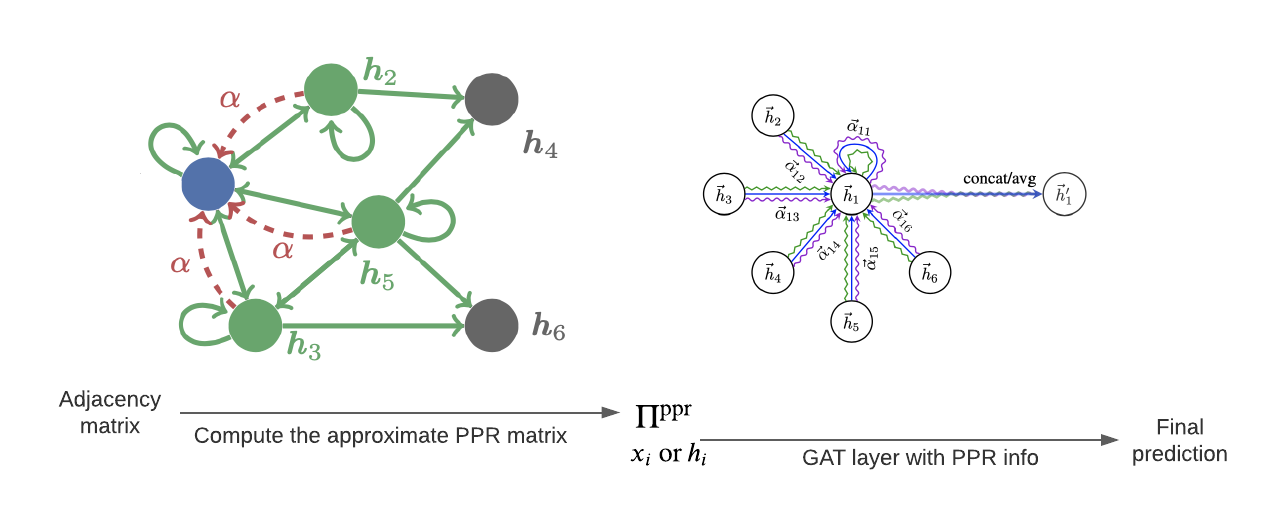}
\end{center}
\caption{Illustration of PPRGAT. First, we precompute $\Pi^{\epsilon, k}$ from the adjacency matrix. Second, $\Pi^{\epsilon, k}_{ij}, x_i, x_j$ are used together to generate the attention score from node $j$ to nose $i$ in the GAT layer. The model is trained end-to-end.}
\label{fig:pprgat}
\end{figure}

Now, when we normalize the attention scores across all neighbors $j \in \mathcal{N}_i$ as in Equation (\ref{eq:gat-attention}) and aggregate the features as in Equation (\ref{eq:gat-agg}), we consider two approaches for the definition of neighbors $\mathcal{N}_i$. First, the base version is to aggregate over the top-k nodes of $\pi(i)$. Second, we aggregate over the original neighbors defined by the original adjacency matrix. We call this variant PPRGAT-local and PPRGATv2-local, respectively. Fig.~\ref{fig:pprgat} shows the overall  illustration of PPRGAT model.


\section{Experiments}
We compare our proposed model against two GAT based baselines and two PPR based baselines. Specifically, GAT based baselines include GAT \cite{gat} and GATv2 \cite{gatv2}, and PPR based baselines include (A)PPNP \cite{appnp} and PPRGo \cite{pprgo}. For our proposed models, we evaluate the four variants described in Section~\ref{sec:method}: PPRGAT, PPRGAT-local, PPRGATv2, and PPRGATv2-local.

\subsection{Datasets}

\textbf{Transductive learning} We test the models on the three standard citation network benchmark datasets: Cora, Citeseer, and Pubmed \cite{sen2008collective}. For these datasets, we follow the transductive experimental setup of \cite{yang2016revisiting}. While large graphs do not necessarily have a larger diameter, note that these graphs indeed have average shortest path lengths between 5 and 10 and therefore a regular two-layer GCN cannot cover the entire graph. This characteristic makes these citation datasets particularly useful to evaluate the effectiveness of introducing PPR information into the shallow layers of GNNs.\\
\textbf{Inductive learning} We test the models on protein-protein interaction (PPI) dataset that consists of graphs corresponding to different human tissues \cite{Zitnik2017}. In this task, the testing graphs remain completely unobserved during training.

Table~\ref{table:datasets} summarizes the graph datasets used in the experiments.


\begin{table*}[]
\caption{Summary of the graph datasets used in the experiments}
\label{table:datasets}
\begin{center}
\begin{tabular}{lrrrr}
\hline
\textbf{}           & \multicolumn{1}{l}{\textbf{Cora}}  & \multicolumn{1}{l}{\textbf{Citeseer}} & \multicolumn{1}{l}{\textbf{Pubmed}} & \textbf{PPI}               \\ \hline
Task                & \multicolumn{1}{l}{Transductive}   & \multicolumn{1}{l}{Transductive}      & \multicolumn{1}{l}{Transductive}    & Inductive                  \\
Nodes               & \multicolumn{1}{l}{2708 (1 graph)} & \multicolumn{1}{l}{3327 (1 graph)}    & \multicolumn{1}{l}{19717 (1 graph)} & 56944 (24 graphs)          \\
Edges               & 5429                               & 4732                                  & 44338                               & \multicolumn{1}{r}{818716} \\
Features/Node       & 1433                               & 3703                                  & 500                                 & \multicolumn{1}{r}{50}     \\
Classes             & 7                                  & 6                                     & 3                                   & 121           \\
Training Examples   & 140                                & 120                                   & 60                                  & 44906          \\
Validation Examples & 500                                & 500                                   & 500                                 & 6514            \\
Test Examples       & 1000                               & 1000                                  & 1000                                & 5524            \\ \hline
\end{tabular}
\end{center}
\end{table*}

\subsection{Experimental Setup}
\label{ssec:setup}
For fair comparisons, for all of our 4 variants of PPRGAT, we use the same hyperparameters as the GAT based baselines and PPR based baselines if possible. Specifically, GAT based setup is for our 4 variants of PPRGAT and the two baselines: GAT and GATv2. PPR based setup is applicable to our 4 variants of PPRGAT and the two baselines: (A)PPNP and PPRGo. More details are described below.

\subsubsection{Transductive learning}

\textbf{Common setup} We use an early stopping strategy on validation loss, with a patience of 100 epochs.\\
\textbf{GAT related setup} We apply a two-layer GAT model. For the first layer, we use  8 features and 8 attention heads. Then it's followed by a ReLU. For the second layer, the number of output features is the same as the number of classes, and we use 1 head except for PubMed. For PubMed, we use 8 output attention heads, following the observation from \cite{monti2016geometric}.\\
\textbf{PPR related setup}
We use teleport parameter $\alpha = 0.25$ for all datasets. For the  approximation $\Pi^{\epsilon, k}$ of $\Pi^{\text{ppr}}$, we use $\epsilon=10^{-4}$ and $k=32$. In other words, we keep only the top 32 elements of each PPR distribution $\pi(i)$.

\subsubsection{Inductive learning}

\textbf{Common setup} We use an early stopping strategy on micro-F1 score on the validation sets, with a patience of 100 epochs.\\
\textbf{GAT related setup} For the inductive learning task (PPI dataset), we apply a three-layer GAT model. Both of the first two layers consist of K = 4 attention heads computing 256 features, followed by an ELU nonlinearity. The final layer is used for the multi-label classification. For this final layer, we use 6 attention heads computing and 121 features each. As observed by \cite{gat}, the training sets for the PPI dataset are sufficiently large, and we found no need to apply dropout for regularization.\\
\textbf{PPR related setup}
We use the same settings as in the transductive learning.

\subsection{Evaluation Results}
The evaluation results of our experiments are summarized in Table \ref{tab:acc} and Table \ref{tab:F1}. The experiment results successfully demonstrate that our models outperform the baselines across all four datasets. It is important to note that our models outperform the baseline on the PPI dataset. This implies that the approximate PPR matrix of the unseen data along side the learned weights can still efficiently predict the outputs.

\begin{table}[t]
\caption{Classification accuracies (in \%) of different node classification algorithms on the citation datasets. Results are the averages of 10 runs.}
\label{tab:acc}
\begin{center}
\begin{tabular}{lrrr}
\hline
\textbf{Model} & \multicolumn{1}{l}{\textbf{Cora}} & \multicolumn{1}{l}{\textbf{Citeseer}} & \multicolumn{1}{l}{\textbf{Pubmed}} \\ \hline
GAT            & 83.0                                & 70.8                                  & 79.0                                  \\
GATv2          & 82.9                              & 71.6                                  & 78.7                                \\
APPNP          & 83.3                              & 71.8                                  & 80.1                                \\
PPRGo          & 74.2                              & 65.6                                  & 70.7                                \\
PPRGAT         & 83.9                              & \textbf{72.5}                         & 80.4                                \\
PPRGAT-local     & \textbf{84.0}                       & 72.2                                  & 80.1                                \\
PPRGATv2       & 83.8                              & 72.4                                  & \textbf{80.5}                       \\
PPRGATv2-local   & 83.9                              & 72.1                                  & 80.2                                \\ \hline
\end{tabular}
\end{center}
\end{table}

\begin{table}[t]
\caption{Classification micro-F1 scores (in \%) of different node classification algorithms on the PPI dataset. Results are the averages of 10 runs.}
\label{tab:F1}
\begin{center}
\begin{tabular}{lr}
\hline
\textbf{Model} & \multicolumn{1}{l}{\textbf{PPI}} \\ \hline
GAT            & 96.5                             \\
GATv2          & 96.3                             \\
APPNP          & 96.7                             \\
PPRGo          & 87.8                             \\
PPRGAT         & \textbf{97.5}                    \\
PPRGAT-local     & 97.1                    \\
PPRGATv2       & 97.4                             \\
PPRGATv2-local   & 97.1                             \\ \hline
\end{tabular}
\end{center}
\end{table}

\section{Conclusion}

We have introduced PPRGAT that incorporates the PPR information into the GATs. In this way, PPRGAT utilizes the expressive power of GATs, while incorporating the whole adjacency matrix information via PPR matrix, even with the shallow GAT layers. Approximating the PPR matrix is very efficient following the approach by \cite{andersen2006local}, and it can be easily parallelized. Furthermore, computing the PPR matrix is one-time preprocess step before starting the training. This PPR matrix is stored in memory and reused during the training. As a result, PPRGATs achieve the scalability at the same level of GATs.

\vfill\pagebreak



\bibliographystyle{IEEEbib}
\bibliography{graph}

\begin{thebibliography}{10}

\bibitem{gori2005new}
Marco Gori, Gabriele Monfardini, and Franco Scarselli,
\newblock ``A new model for learning in graph domains,''
\newblock in {\em Proceedings. IEEE International Joint Conference on Neural
  Networks}. IEEE, 2005, vol.~2, pp. 729--734.

\bibitem{scarselli2008graph}
Franco Scarselli, Marco Gori, Ah~Chung Tsoi, Markus Hagenbuchner, and Gabriele
  Monfardini,
\newblock ``The graph neural network model,''
\newblock {\em IEEE transactions on neural networks}, vol. 20, no. 1, pp.
  61--80, 2008.

\bibitem{gilmer2017neural}
Justin Gilmer, Samuel~S Schoenholz, Patrick~F Riley, Oriol Vinyals, and
  George~E Dahl,
\newblock ``Neural message passing for quantum chemistry,''
\newblock in {\em International conference on machine learning}. PMLR, 2017,
  pp. 1263--1272.

\bibitem{kearnes2016molecular}
Steven Kearnes, Kevin McCloskey, Marc Berndl, Vijay Pande, and Patrick Riley,
\newblock ``Molecular graph convolutions: moving beyond fingerprints,''
\newblock {\em Journal of computer-aided molecular design}, vol. 30, no. 8, pp.
  595--608, 2016.

\bibitem{gat}
Petar Veličković, Guillem Cucurull, Arantxa Casanova, Adriana Romero, Pietro
  Liò, and Yoshua Bengio,
\newblock ``Graph attention networks,''
\newblock in {\em International Conference on Learning Representations}, 2018.

\bibitem{pinsage}
Rex Ying, Ruining He, Kaifeng Chen, Pong Eksombatchai, William~L. Hamilton, and
  Jure Leskovec,
\newblock ``Graph convolutional neural networks for web-scale recommender
  systems,''
\newblock in {\em Proceedings of the 24th ACM SIGKDD International Conference
  on Knowledge Discovery and Data Mining}. 2018, p. 974–983, Association for
  Computing Machinery.

\bibitem{li2018deeper}
Qimai Li, Zhichao Han, and Xiao-Ming Wu,
\newblock ``Deeper insights into graph convolutional networks for
  semi-supervised learning,''
\newblock in {\em Thirty-Second AAAI conference on artificial intelligence},
  2018.

\bibitem{fastgcn}
Jie Chen, Tengfei Ma, and Cao Xiao,
\newblock ``Fastgcn: Fast learning with graph convolutional networks via
  importance sampling,'' 2018.

\bibitem{page1999pagerank}
Lawrence Page, Sergey Brin, Rajeev Motwani, and Terry Winograd,
\newblock ``The pagerank citation ranking: Bringing order to the web.,''
\newblock Tech. {R}ep., Stanford InfoLab, 1999.

\bibitem{appnp}
Johannes Klicpera, Aleksandar Bojchevski, and Stephan G{\"u}nnemann,
\newblock ``Predict then propagate: Graph neural networks meet personalized
  pagerank,''
\newblock in {\em International Conference on Learning Representations (ICLR)},
  2019.

\bibitem{pprgo}
Aleksandar Bojchevski, Johannes Klicpera, Bryan Perozzi, Amol Kapoor, Martin
  Blais, Benedek R{\'o}zemberczki, Michal Lukasik, and Stephan G{\"u}nnemann,
\newblock ``Scaling graph neural networks with approximate pagerank,''
\newblock in {\em Proceedings of the 26th ACM SIGKDD International Conference
  on Knowledge Discovery and Data Mining}, New York, NY, USA, 2020, ACM.

\bibitem{kipf2017semi}
Thomas~N. Kipf and Max Welling,
\newblock ``Semi-supervised classification with graph convolutional networks,''
\newblock in {\em International Conference on Learning Representations (ICLR)},
  2017.

\bibitem{graphsage}
William~L. Hamilton, Rex Ying, and Jure Leskovec,
\newblock ``Inductive representation learning on large graphs,'' 2018.

\bibitem{xu2015show}
Kelvin Xu, Jimmy Ba, Ryan Kiros, Kyunghyun Cho, Aaron Courville, Ruslan
  Salakhudinov, Rich Zemel, and Yoshua Bengio,
\newblock ``Show, attend and tell: Neural image caption generation with visual
  attention,''
\newblock in {\em International conference on machine learning}. PMLR, 2015,
  pp. 2048--2057.

\bibitem{gatv2}
Shaked Brody, Uri Alon, and Eran Yahav,
\newblock ``How attentive are graph attention networks?,''
\newblock in {\em International Conference on Learning Representations}, 2022.

\bibitem{andersen2006local}
Reid Andersen, Fan Chung, and Kevin Lang,
\newblock ``Local graph partitioning using pagerank vectors,''
\newblock in {\em IEEE Symposium on Foundations of Computer Science (FOCS'06)}.
  IEEE, 2006, pp. 475--486.

\bibitem{nassar2015strong}
Huda Nassar, Kyle Kloster, and David~F Gleich,
\newblock ``Strong localization in personalized pagerank vectors,''
\newblock in {\em International Workshop on Algorithms and Models for the
  Web-Graph}. Springer, 2015, pp. 190--202.

\bibitem{sen2008collective}
Prithviraj Sen, Galileo Namata, Mustafa Bilgic, Lise Getoor, Brian Galligher,
  and Tina Eliassi-Rad,
\newblock ``Collective classification in network data,''
\newblock {\em AI magazine}, vol. 29, no. 3, pp. 93--93, 2008.

\bibitem{yang2016revisiting}
Zhilin Yang, William Cohen, and Ruslan Salakhudinov,
\newblock ``Revisiting semi-supervised learning with graph embeddings,''
\newblock in {\em International conference on machine learning}. PMLR, 2016,
  pp. 40--48.

\bibitem{Zitnik2017}
Marinka Zitnik and Jure Leskovec,
\newblock ``Predicting multicellular function through multi-layer tissue
  networks,''
\newblock {\em Bioinformatics}, vol. 33, no. 14, pp. 190--198, 2017.

\bibitem{monti2016geometric}
F~Monti, D~Boscaini, J~Masci, E~Rodola, J~Svoboda, and MM~Bronstein,
\newblock ``Geometric deep learning on graphs and manifolds using mixture model
  cnns,''
\newblock {\em IEEE conference on computer vision and pattern recognition},
  2016.

\end{thebibliography}

\end{document}